\documentclass[accepted]{uai2021} 

\usepackage[american]{babel}

\usepackage{natbib} 
\usepackage{mathtools} 
\usepackage{booktabs} 
\usepackage{tikz} 

\usepackage{epsfig}
\usepackage{amsmath}
\usepackage{amssymb}
\usepackage{cuted}
\usepackage{flushend}
\usepackage{textcomp}
\usepackage{subcaption}
\usepackage{color}
\usepackage{comment}

\let\emptyset\varnothing

\DeclareMathOperator*{\argmin}{arg\,min}
\DeclareMathAlphabet{\mathcalold}{OT1}{pzc}{m}{it}

\usepackage{color}

\title{An Unsupervised Video Game Playstyle Metric via State Discretization}

\author[1]{Chiu-Chou~Lin\thanks{dsobscure@outlook.com}}
\author[1,2]{Wei-Chen~Chiu\thanks{walon@cs.nctu.edu.tw}}
\author[1,2,3]{I-Chen~Wu\thanks{icwu@cs.nctu.edu.tw}}
\affil[1]{%
    Department of Computer Science, National Yang Ming Chiao Tung University, Taiwan
}
\affil[2]{%
    Pervasive Artificial Intelligence Research (PAIR) Labs, Taiwan
}
\affil[3]{%
    Research Center for IT Innovation, Academia Sinica, Taiwan
}
  
\begin{document}
\maketitle

\begin{abstract}
On playing video games, different players usually have their own playstyles. 
Recently, there have been great improvements for the video game AIs on the playing strength. 
However, past researches for analyzing the behaviors of players still used heuristic rules or the behavior features with the game-environment support, thus being exhausted for the developers to define the features of discriminating various playstyles.
In this paper, we propose the first metric for video game playstyles directly from the game observations and actions, without any prior specification on the playstyle in the target game.
Our proposed method is built upon a novel scheme of learning discrete representations that can map game observations into latent discrete states, such that playstyles can be exhibited from these discrete states. 
Namely, we measure the playstyle distance based on game observations aligned to the same states. 
We demonstrate high playstyle accuracy of our metric in experiments on some video game platforms, including TORCS, RGSK, and seven Atari games, and for different agents including rule-based AI bots, learning-based AI bots, and human players. 
\end{abstract}

\section{Introduction}
\label{Introduction}
Generally, players of a video game would demonstrate different playstyles~\citep{Defining-personas,Playstyle-Evolve} to enrich playing experiences.
Take a racing game as an example: different players usually have their own preferences on the track positions or the strategies of using accelerations.
Basically, the playstyle can be captured by the characteristic of the sets of playing behaviors, where the playing behavior is represented by the pair of game observation (i.e., screens) and its corresponding action taken by the player.

Although it is known that the player modelling has been a common domain in studying the behaviors of players~\citep{player_modeling}, the playstyle which we are interested here is able to deliver the intention or preferences of agents or players.
If we can directly measure playstyles rather than just capturing some playing behaviors, it becomes useful to develop AI bots for following various playstyles, such as mimicking human players' styles. 
Therefore, how to measure playstyles becomes important with great potential to both the video game and the AI communities.

Recently, with the rapid growth of deep reinforcement learning (DRL) techniques, the trained AI bots not only achieve super-human performance in several complicated games, e.g., DotA2~\citep{open_five} and StartCraft II~\citep{alpha_star}, but also attempt to discover new playstyles during their training procedure~\citep{hide_and_seek}.
However, it is hard to have proper metrics to measure the diversity among the playing polices of different AI bots, hence being difficult for game developers to define a specific playstyle or even enrich the playstyles in a new video game.
To the best of our knowledge, the existing works for the player behaviors still mainly consider the heuristic rules or study behavior features based on the explicit game-environment support \citep{Defining-personas,Playstyle-Evolve,Playing-style-recognition,player_modeling,cluster_player_behaviour}, where they are limited to some pre-defined playstyles. 
So, it is hard to support a general playstyle metric in the above way. 
It hence becomes more expected to have a general metric to measure playstyles without any prior knowledge about styles or game specifications.

The issue of designing a general metric for playstyles is actually nontrivial in the sense that different combinations of playing behaviors may lead to too many different game results to make it hard to identify playstyles. 
In the past, researches related to playstyles were studied on some applications, such as the categorization of driving behaviors in the driver assistance systems \citep{driving-style-nn,topological_map}.
Some of these researches discovered different driving behaviors by using discrete representations of the states from sensor signals (derived by Hidden Markov Models (HMMs) \citep{HMM}), and then did clustering for driving styles afterwards.
Many works followed to utilize these representations to perform analysis of the time-series data of driving \citep{DAA, DAA-TP,NPYLM}.
As an extension of the analysis of driving behaviors, \citet{topological_map} propose a metric for evaluating the similarity between driving styles based on the patterns derived from the t-distributed stochastic neighbor embedding (t-SNE) of image observations.

In addition to analyzing driving behaviors, another possible solution for evaluating the similarity between playstyles from observations is to estimate the distance between distributions of image observations (i.e., game screens) produced during the game playing, via the metric such as Fr\'{e}chet Inception Distance~\citep{FID}. 
Nevertheless, the above solution does not take actions into account. 
Besides, some playstyles depend on random events in a game, which does not exist in common image distribution comparisons.
Therefore, there exists a significant demand for the metric capable of directly predicting playstyles from the rich information composed of the observation-action pairs.
This is an issue to be addressed in this paper. 

In this paper, we propose a novel metric to predict the video game playstyles based on observations and actions of game episodes without any prior knowledge about style specifications. 
Our proposed method, inspired by the above driving styles, is built upon a novel scheme of learning discrete representations, called hierarchical state discretization (HSD) in this paper, which can map game observations into latent discrete states, such that playstyles can be exhibited from these discrete states.
Discrete representation is reviewed in Section \ref{Background}, and HSD is described in Section \ref{Hierarchical State Discretization}. 
Namely, we measure the playstyle distance based on game observations aligned to the same states as presented in Section \ref{General Playstyle Metric}. 
Then, we predict playstyles based on the playstyle distance. 
In our experiments described in Section \ref{Experiments}, we demonstrate high playstyle accuracy of our proposed metric on three video game platforms, including TORCS~\citep{TORCS}, RGSK\footnote{\url{https://assetstore.unity.com/packages/templates/racing-game-starter-kit-22615}}, and seven Atari games~\citep{atari}, and for different agents including rule-based AI bots, learning-based AI bots, and human players. 
Finally, we make concluding remarks in Section \ref{Conclusion}. 


\section{Background}
\label{Background}
The playstyle is closely related to the mapping from the states of the game environment (i.e., the visual observations or game screens in a video game) to the actions taken by the agent/player (which actually being analogous to the \textit{policy}). 
Hence, the metric of playstyle intuitively would like to compare if two agents intend to take similar actions with respect to (nearly) the same states. 
However, when the observation space is high-dimensional or continuous, it is typically hard to find the same observations across different game recordings as the basis for further comparing the actions.
In order to deal with such a problem, we introduce a method that learns the discrete representations of the observation space, such that the probability of having the same states across recordings significantly increases. 
A naive solution for reducing the observation space could be from the simple downsampling strategy, for instance, downsampling a given game screen into an image with a much smaller resolution. 
However, even when we have a low-resolution image after performing downsampling, e.g., 8$\times$8 image size, with assuming there are only 16 possible intensity values for each pixel, the state space will end up with the size of $16^{64}$, which is still too large to be feasible for our performing the further metric computation.
Another well-known solution for obtaining discrete representations comes from the Vector Quantised-Variational AutoEncoder (VQ-VAE), proposed by \citep{VQ-VAE}, where a 
well-configured VQ-VAE model can extract latent discrete states with contextual information in video games, in which it fits the needs for our playstyle metric.

\begin{figure*}
    \begin{center}
        \includegraphics[width=1\textwidth]{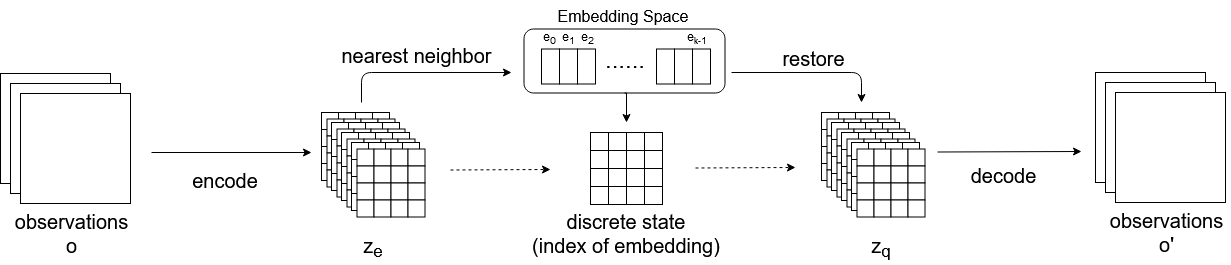}
    \end{center}
    \caption{Illustration of  VQ-VAE architecture~\citep{VQ-VAE}}
    \label{Figure:VQ-VAE}
\end{figure*}

A typical VQ-VAE model has three components: encoder, decoder, and the embedding space, as shown in Figure~\ref{Figure:VQ-VAE}. 
In a VQ-VAE model, the encoder first projects the image observation $o$ to a latent feature map $z_e$. Then, the nearest neighbor method with respect to $K$ shared tensors $E=\{e_1, \cdots, e_K\}$ is used to turn $z_e$ into a discrete representation $\overline{s}$, that is, we replace each cell $z_e^i$ of the $z_e$ feature map with its nearest neighbor in $E$ by the following equation.
\begin{equation}
\label{equation:VQ-VAE}
q(\overline{s}=k|o)=
\begin{cases}
    1 & \quad \text{for } k_{i} = \underset{j}{\argmin} ||z_{e}^{i}(o)-e_{j}||^2 \text{,}\\
    0 & \quad \text{otherwise.}
\end{cases}
\end{equation}
Basically, the discrete state $\overline{s}$ is a ordered group of indexes to the elements in $E$, in which $\overline{s}$ can be easily recovered back to the continuous form $z_q$ by creating a feature map (having same size as $z_e$) with its cells coming from the corresponding tensors in $E$ of each index in $\overline{s}$.   
Finally, the decoder attempts to reconstruct the observation $o$ from the restored latent code $z_q$.
Derived from VQ-VAE, we develop a novel extension to further reduce the latent discrete space  for enabling the computation of our playstyle metric in Section~\ref{Hierarchical State Discretization}.

\section{Playstyle Metric}
\label{General Playstyle Metric}
This section presents our definition for playstyle metrics.
In Section \ref{Metric:datasets}, we describe the terminology of the playing dataset and playstyle. 
Then, we define the playstyle distance in Section \ref{Metric:policy_distance} and discuss its properties in Section \ref{Metric:Discussion}. 

\subsection{Representations of Playstyles}
\label{Metric:datasets}

In this subsection, we present the representation of the playstyles. 
An agent plays a set of game episodes, which are recorded as game recordings and collected as playing datasets for 
playstyle evaluation.
In reality, an agent may play in several different styles in these game episodes.
For example, an agent tends to drive at the speed of 60 in a segment of a game episode but at the speed of 90 in some other segment.
However, for simplicity of analysis, in this paper, we assume that an agent always plays in a single playstyle in a set of game recordings, which are collected as a playing dataset; that is, {\em a playing dataset represents one playstyle}. 
So, some datasets collected from a single agent are supposed to have the same playstyle or have a high similarity in playstyle. 
For example, let an AI-bot tend to play at the speed of 60. Thus, all datasets collected from the bot should have high similarity. 

A playstyle metric is to measure the similarity of two playstyles. 
In this paper, we measure the playstyle distance between two datasets (representing two playstyles), which will be described in the next subsection instead.
If these two playstyles have a small playstyle distance, we say that the two corresponding playstyles tend to have high similarity.

\subsection{Playstyle Distance}
\label{Metric:policy_distance}
Our metric aims to measure how close two playing datasets $A$ and $B$ are. 
A dataset, say $A$, is a set of pairs of observations and their corresponding actions in the action space $\mathcalold{A}$. 
Let $O_{A}$ denotes its observation set, a set of observations.\footnote{An observation is usually represented by a tensor in the way of how a player observes the game, e.g., consecutive 4 images of the game screen is a common observation in the video games~\citep{dqn}.} 
For an observation $o$, an action $a \in \mathcalold{A}$ is associated with $o$ to indicate how the agent acts to $o$. 
If the action space is discrete (e.g., the keyboard inputs), we can use the categorical distribution for modeling the action distribution. 
If the action space is continuous (e.g., inputs from a scroll wheel), we turn to use the multivariate normal distribution. 
In our metric, an observation $o$ is mapped to a discrete state $\overline{s}$ via a mapping function $\phi$ from an observation set $O$ to a set of states $\overline{S}$. 
In this paper, a mapping function based on state discretization is described in Section \ref{Hierarchical State Discretization}.

In order to compare the action distributions between two datasets, say $A$ and $B$, we first derive the intersection $\overline{S}_\phi$ of the discrete states from $O_{A}$ and $O_{B}$ respectively, where
\begin{equation}
\label{equation:intersection_state_space}
\overline{S}_\phi(A,B) = \phi(O_{A})\cap\phi(O_{B}).
\end{equation}
Given a state $\overline{s} \in \overline{S}_\phi$ and its corresponding action distributions $\pi_{A}$ and $\pi_{B}$ from $A$ and $B$ respectively,
define a policy distance on $\overline{s}$ as 
\begin{equation}
\label{equation:conditional_policy_distance}
\overline{d}_\phi(A,B|\overline{s}) = W_{2}(\pi_A, \pi_B|\overline{s}),
\end{equation}
where 2-Wasserstein distance ($W_2$) ~\citep{first_define_w2,W2-Metric} is used to calculate the distance between distributions in this paper, as it is widely used in the metric of generative models, e.g., FID~\citep{FID}. 
The idea of using this way is to compare actions on those observations mapped to the same states only. 

This paper presents two methods to calculate the playstyle distance between two datasets $A$ and $B$ with the mapping function $\phi$.
The first is to average the distance uniformly over all the intersected states as 
\begin{equation}
\label{equation:policy_distance_uniform}
d_{\phi}(A,B) 
= \frac{
    \sum_{\overline{s}\in\overline{S}_\phi(A,B)} \overline{d}_\phi(A,B|\overline{s}) }
{|\overline{S}_\phi(A,B)|}, 
\end{equation}
where every state $\overline{s}$ in $\overline{S}_\phi(A,B)$ has the same importance for the playstyle metric. 
Obviously, it is symmetric in the sense of $d_{\phi}(A,B)=d_{\phi}(B,A)$.
The second is to calculate the expected distance according to the observation distribution of $O_{A}$ and $O_{B}$ over $\overline{S}_\phi$, defined as
\begin{equation}
\label{equation:style_distance}
d_{\phi}(A,B) = \frac{d_{\phi}(A|B)}{2} + \frac{d_{\phi}(B|A)}{2}\text{,}
\end{equation}
where 
\begin{equation}
\label{equation:cond_style_distance}
d_{\phi}(X|Y) = \mathbb{E}_{o \sim O_Y,\phi(o) \in \overline{S}_\phi(X,Y)}[\overline{d}_\phi(X,Y|\phi(o))].
\end{equation}
This method is symmetric too, but the expected distance weighs more for those with high observation occurrences in the game playing.
To distinguish both methods, let us illustrate the following case for two states: first, many observation occurrences on both $A$ and $B$, say $n_A$ observation occurrences in $A$ and $n_B$ observation occurrences in $B$, are mapped to a state, say $s_1\in\overline{S}_\phi$; second, only one observation occurrence on each of $A$ and $B$ are mapped to a set, say $s_2\in\overline{S}_\phi$.
In this case, the policy distance for $s_1$ weighs $n_A$ times more for $d_{\phi}(A|B)$ (respectively $n_B$ times more for $d_{\phi}(B|A)$) in the second method than the first, while the policy distance for $s_1$ weighs one for both methods.
An example in more detail is given in the appendix. 
For the above two, if $\overline{S}_\phi=\emptyset$, the playstyle distance is set to undefined which will be further described in the section for experiments, as there is no common states across two playing datasets to measure their playstyle similarity.
For simplicity of discussion, the rest of this paper are based on the second only, since the experiment results described in the appendix shows the second performs better than the first. 

In practical implementation, we add an additional condition to $\overline{S}_\phi$ with a threshold $t$ to discard the rarely visited states that tends to be unstable during the metric computation. 
With the threshold $t$, we use the following, instead of $\overline{S}_\phi$. 
\begin{equation}
\label{equation:intersection_state_with_threshold}
\begin{aligned}
& \overline{S'}_\phi(A,B,t) = \{\overline{s} | \overline{s} \in \overline{S}_\phi(A,B),  
F_\phi(\overline{s},A) \geq t, 
F_\phi(\overline{s},B) \geq t\}\text{,}\\
& \text{where } F_\phi(s,X) = |\{o | o \in O_{X}, \phi(o) \in \{s\}\}|.
\end{aligned}
\end{equation}

\subsection{Discussion}
\label{Metric:Discussion}

As motivated previously in Section \ref{Introduction}, prior works on modeling player behaviors usually require domain-specific knowledge to extract relevant features or define target behaviors, which is hard to be generalized to other video games. 
Our playstyle metric in the previous subsection addresses the following two issues. 
The first one is \textit{generality}. 
In our metric, we have no assumption on the targeting playstyle label, except for states and actions. 
In other word, the metric is capable of differentiating several playstyles in a game from various playing datasets, such that the metric does not need any predefined or specific features to the targeted playstyles, such as the playing speeds, 
or positions for racing cars.

The second one is \textit{consistency}.
As described in the previous subsections, two datasets with the same playstyle (or from the same agent) tends to have small playstyle distance in comparison to those with different playstyles, in terms of the metric.
While playstyle similarity or distance is subjective in some sense, we assume that there exists an ideal or oracle metric to measure playstyle distance. 
A metric of estimating playstyle distance $d^*(A,B)$ is said to be consistent to an ideal similarity metric, if the following is satisfied. 
\begin{description}
\item[Consistency] Given any three playing datasets $A$, $B$ and $C$, the similarity between $A$ and $C$ is higher than that between $B$ and $C$ in terms of the oracle metric, if and only if $d^*(A,C) < d^*(B,C)$.
\end{description}
In this paper, we argue that the expected playstyle distance in (\ref{equation:cond_style_distance}) in our metric distinguishes the distance relatively after a sufficient large number of samples. 
Namely, we argue that the following property is satisfied for any three playing datasets $A$, $B$ and $C$, where the similarity between $A$ and $C$ is higher than that between $B$ and $C$. 
\begin{description}
\item[Argument] 
Given a sufficient large number of samples, the expected playstyle distance in (\ref{equation:cond_style_distance}) satisfies the following:
$d_{\phi}(A|C) < d_{\phi}(B|C)$ and $d_{\phi}(C|A) <d_{\phi}(C|B)$.
\end{description}

The argument $d_{\phi}(A|C)<d_{\phi}(B|C)$ implies that the expected distance from $C$ to $A$ is closer than that from $C$ to $B$ according to $C$'s observation distribution. 
Next, the argument  $d_{\phi}(C|A)<d_{\phi}(C|B)$ is similar, but in a different aspect where the observation distribution is reversed, namely, the targeting datasets $A$ and $B$ use their own observation distribution to calculate the expected distance to $C$. 
We can consider the targeting datasets $A$ and $B$ have consensus on the relation of distance value. 
With these arguments, we derive that after a sufficient large number of samples the playstyle distance given in (\ref{equation:cond_style_distance}) is consistent as follows. 
\begin{equation}
\begin{aligned}
& d_{\phi}(A,C) & = & \quad d_{\phi}(A|C)/2 + d_{\phi}(C|A)/2 \\ 
& \quad & < & \quad d_{\phi}(B|C)/2 + d_{\phi}(C|B)/2 \\
& \quad & = & \quad d_{\phi}(B,C) 
\end{aligned}
\end{equation}

\section{Hierarchical State Discretization}
\label{Hierarchical State Discretization}
As described in Section~\ref{Metric:policy_distance}, we need a proper discrete state mapping function $\phi$ to find the intersection states between playing datasets.
Thus, we develop a novel discrete representation learning method for enabling our proposed playstyle metric in the high dimensional observation space. 
To start with, we first provide the description on the design of our discrete representation learning in Subsection~\ref{HSD:Architecture-Design}.
Next, we introduce the gradient copy trick in VQ-VAE in Subsection~\ref{HSD:Update-VQ-VAE-Model}.
Finally, the training procedure of our model is defined in Subsection~\ref{HSD:Update-HSD-Model}.

\subsection{Architecture Design}
\label{HSD:Architecture-Design}
\begin{figure*}
    \begin{center}
        \includegraphics[width=1\textwidth]{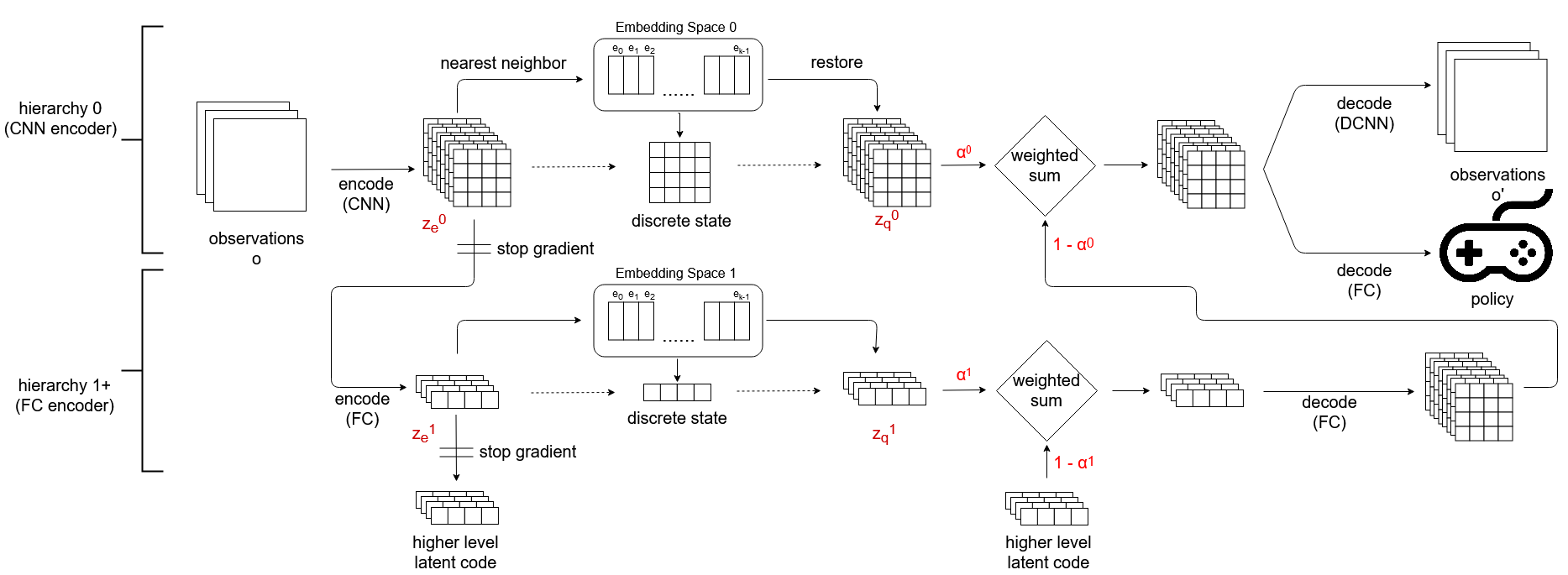}
    \end{center}
    \caption{Architecture of our proposed hierarchical state discretization (HSD) model}
    \label{Figure:HSD}
\end{figure*}

A well-configured VQ-VAE model is able to extract the contextual information from raw observations in the video games~\citep{VQ-VAE}; however, the size of state space in a typical VQ-VAE model could be still too large for our playstyle metric computation.
To better control the size of state space, while keeping the capacity of the latent representation during the model optimization, we propose a multi-hierarchy stacked structure of the VQ-VAE model, named as \textit{Hierarchical State Discretization} (\textbf{HSD}),
where its model architecture is shown in Figure~\ref{Figure:HSD}.

The convolutional encoder of our HSD in the base hierarchy (hierarchy 0, closest to the original observation space) follows the original design in a VQ-VAE model, while we add extra fully-connected encoders above the base hierarchy for controlling the discrete state space. Basically, a fully-connected encoder (in higher hierarchy other than the base one) takes the latent representation obtained by the previous hierarchy as its input and outputs a latent code, which is then folded to the representation composed of
$B$ cells of 1-D feature map
and gone through the vector quantization procedure afterward as VQ-VAE does.
The size of the discrete state space during such procedure is $K^B$, where $K$ is the number of candidate tensors used for quantization in the embedding space at this hierarchy.

In the base hierarchy, in addition to the typical decoder taking care of the observations, we have another decoder related to the policy for encouraging discrete representations to better capture the information of actions, which is also quite important for our playstyle metric computation.
Every hierarchy above the base one has a fully-connected decoder after the vector quantization procedure, which reconstructs the latent representations in the lower hierarchy.
Moreover, we propose a weighted sum block to combine the latent representation of hierarchy $h$ and the decoder output of hierarchy $h+1$ as the input for the decoder of hierarchy $h$, where the coefficients for such weighted sum are $\alpha^{h}$ (sampled from uniform distribution between $0$ and $1$) and $1-\alpha^{h}$ respectively, as shown in Figure~\ref{Figure:HSD}. Such mechanism of using weighted sum blocks helps encouraging the decoder of the base hierarchy to utilize the outputs from all the hierarchies, thus the model training (i.e. gradients propagated from the objective in the base hierarchy) can smoothly contribute to all the subnetworks (i.e. decoders and encoders) in our HSD.
For learning our HSD model, in order to enable the gradient propagation stopped by the nearest neighbor method of vector quantization, we follow the gradient copy trick proposed in VQ-VAE and extend it for our HSD architecture, which will be detailed in the next subsection.

It is worth noting that, there exists another well-known hierarchical VQ-VAE, the VQ-VAE-2 framework proposed by~\citep{VQ-VAE-2}, which is however quite different from our HSD in two perspectives: (1) VQ-VAE-2 is aiming to generate high-quality images with the autoregressive prior in the latent representation, while our HSD instead attempts to discover the intersection states between playing datasets; (2) VQ-VAE-2 jointly uses the discrete latent maps obtained from multiple hierarchies to generate the output image, thus its main purpose is to \textit{increase} the latent states for producing delicate image details, while our HSD is proposed to \textit{reduce} the number of latent states. Moreover, the architectures of VQ-VAE-2 and our HSD are distinct from each other.

\subsection{Gradient Copy Trick}
\label{HSD:Update-VQ-VAE-Model}
There are two losses in gradient computation of a VQ-VAE model:
$L_{rec}$ is the reconstruction error between the observations $o$ and their corresponding reconstruction $o'$, and 
$L_{vq}$ is the vector quantization (VQ) loss between the latent representation $z_e$ and its nearest code $z_q$ in the embedding space (cf. Figure~\ref{Figure:VQ-VAE}). 
We adopt the Huber function~\citep{huber-loss} $D$ for computing the reconstruction error $D(o,o')$ as well as the VQ loss $D(z_e,z_q)$:
\begin{equation}
\label{equation:vq-vae-loss}
L_{rec} = \mathbb{E}[D(o,o')], \quad
L_{vq} = \mathbb{E}[D(z_e,z_q)]
\end{equation}
Let $\theta_{encoder}$ be the weights of the encoder in a VQ-VAE model. 
The gradients of $\theta_{encoder}$ are computed by the chain rule as Equation~\ref{equation:vq-vae-encoder-gradient} with a gradient copy trick (i.e. the gradients are copied from decoder input $z_q$ to encoder output $z_e$~\citep{VQ-VAE}) to make the backpropagation going through the nearest neighbor path, where
$\beta$ is a coefficient to control the updating speed of the embedding vectors, with a default value set to 0.25 suggested by VQ-VAE.
\begin{equation}
\label{equation:vq-vae-encoder-gradient}
\nabla\theta_{encoder} = \frac{\partial L_{rec}}{\partial z_q} \times \frac{\partial z_e}{\partial \theta_{encoder}}  + \beta \times \frac{\partial L_{vq}}{\partial \theta_{encoder}}
\end{equation}

\subsection{Training HSD Model}
\label{HSD:Update-HSD-Model}
Here we introduce the objective functions for training our proposed hierarchical state dicretization (HSD) framework. First, we adopt the same reconstruction loss $L_{rec}$ as original VQ-VAE model on the observations.
Second, for further encouraging the features extracted from our HSD is also aware of the information of policy/action, we add an extra loss function $L_{\pi}$ defined as follows. 
\begin{equation}
\label{equation:hsd-policy-loss}
L_{\pi}=
\begin{cases}
\mathbb{E}[H(\pi_{label},\pi_{decoder})] & \text{if } \mathcalold{A} \text{ is discrete,} \\
\mathbb{E}[D(\pi_{label},\pi_{decoder})] & \text{if } \mathcalold{A} \text{ is continuous,}
\end{cases}
\end{equation}
where $H$ stands for the cross entropy, $\pi_{decoder}$ and $\pi_{label}$  are the action prediction produced by the policy decoder of HSD and the corresponding groundtruth label, respectively. Last, we revise the vector quantization loss of VQ-VAE to a hierarchical fashion $L_{vq}^{h}$ with a hierarchy index $h$ as follows to better fit our HSD framework:
\begin{equation}
\label{equation:hsd-vq-loss}
L_{vq}^{h} = \mathbb{E}[D(z_e^h,z_q^h)]
\end{equation}
As mentioned previously, the decoder in the base hierarchy (i.e., hierarchy index $0$) consists of the observation decoder and the policy decoder, where their weights are $\theta_{rec}$ and $\theta_{\pi}$ respectively. 
The weights of the encoders, the embedding, and the decoders with the hierarchy index $h$ are then denoted as $\theta_{enc}^h$, $\theta_{embed}^h$, and $\theta_{dec}^h$ respectively.
The gradients of each component are computed in Equation~\ref{equation:hsd-encoder-gradient}-Equation~\ref{equation:hsd-other-gradients}.
\begin{equation}
\label{equation:hsd-encoder-gradient}
\begin{aligned}
\nabla\theta_{enc}^h = & \frac{\partial L_{rec}}{\partial z_q^h} \times
\frac{\partial z_e^h}{\partial \theta_{enc}^h} +
\frac{\partial L_{\pi}}{\partial z_q^h} \times
\frac{\partial z_e^h}{\partial \theta_{enc}^h} + \beta \times \frac{\partial L_{vq}^h}{\partial \theta_{enc}^h}
\end{aligned}
\end{equation}
\begin{equation}
\label{equation:hsd-other-gradients}
\begin{aligned}
&\nabla\theta_{embed}^h = \frac{\partial L_{vq}^h}{\partial \theta_{embed}^h}
\text{, }\quad 
\nabla\theta_{dec}^h = \frac{\partial L_{rec}}{\partial \theta_{dec}}  + \frac{\partial L_{\pi}}{\partial \theta_{dec}}
\text{, } \\
&\nabla\theta_{rec} = \frac{\partial L_{rec}}{\partial \theta_{rec} }
\text{, }\quad 
\nabla\theta_{\pi} = \frac{\partial L_{\pi}}{\partial \theta_{\pi}}
\end{aligned}
\end{equation}
where the gradient copy trick is also adopted for the gradient computation of $\theta_{enc}^h$ .

\section{Experiments}
\label{Experiments}

\begin{table}
    \centering
    \begin{tabular}{lllll}
    \toprule
    \quad & Pixel & LRD & HSD & Continuous\\
    \midrule
    25 Styles & 4.00  & 4.00 & \textbf{69.73} & time out \\
    \midrule
    5 Speed Styles & 20.00 & 20.00 & \textbf{91.20} & 27.60\\
    \midrule
    5 Noise Styles & 20.00 & 20.00 & \textbf{81.33} & 23.20\\
    \bottomrule
    \end{tabular}
    \caption{TORCS Playstyle Accuracy (\%). Please refer to Section~\ref{Style On Rule-based AI Bot: TORCS} for detailed description.}
    \label{Table:TORCS-Style-Accuracy-Baseline}
\end{table}

This subsection evaluates our proposed playstyle metric on several video game platforms, including TORCS, RGSK, and seven Atari games.
As described in Section \ref{General Playstyle Metric},  we assume that a playing dataset represents one playstyle and that an agent always plays in a single playstyle in a dataset. 
So, in our experiment, let different agents $p$ produce their own datasets $C_p$ as candidates, whose collection is denoted by $\mathcalold{C}=\{C_p\}$. 
Then, let these agents produce additional datasets  $T_p$ as targets. 
For $T_p$, find the most matching playstyle $C$ among the collection $\mathcalold{C}$ in the following formula. 
\begin{equation}
\label{experiment:retrieval_formula}
C = \underset{C' \in \mathcalold{C}}{\argmin} \text{ } d_\phi(T_p,C').
\end{equation}
If $C$ is $C_p$, also played by the agent $p$, then it is said to be a correct prediction. 
Then, the \textit{playstyle prediction accuracy} is defined as the ratio of correct predictions over the total number of trials, namely 100 trials for all $T_p$ in our experiments. 
The accuracy is used to evaluate the performance of the playstyle metric.

In Subsection~\ref{Style On Rule-based AI Bot: TORCS}, we analyze the playstyles of rule-based AI players in the racing game, TORCS. 
Then, in Subsection~\ref{Style On Human Player: RGSK}, we recruit human players to play the racing game, RGSK, for evaluating the metric upon the practical playstyles.  
Finally, we adopt several learning-based AI players to play different Atari games in Subsection~\ref{Style On Learning-based AI Bot: Atari2600} to study the efficacy of our proposed metric on a wide range of video games.

\subsection{TORCS}
\label{Style On Rule-based AI Bot: TORCS}
In this subsection, our experiment is to analyze the playstyles of the racing game, TORCS~\citep{TORCS}, by the rule-based AI agents developed in \citep{gym-torcs}, which are able to control the acceleration and steering angle of the race car to maintain a target speed in a given default track. 
Besides, since the randomness in the single-player mode of TORCS is low, we also inject noises sampled from a normal distribution into the action values for generating diverse games. 

Now, let us produce playing datasets $C_{s,n}$ as candidates, which contains a set of samples, namely 1024 samples in our experiment, from game episodes played by the above rule-based agent with the speed $s$ and a noise level $n$.
From Section \ref{General Playstyle Metric}, each $C_{s,n}$ represents a playstyle. 
In this experiment, we use 5 different target speeds (i.e. 60, 65, 70, 75, and 80) and 5 noise levels (denoted by $n1$, $n2$, $n3$, $n4$ and $n5$) for the action value.
For each noise level, the action noises are sampled from two zero-mean Gaussian distribution related to the steers and accelerations; and the standard deviation (of the steers and accelerations respectively) for 5 noise levels are (0.01,0.005), (0.02,0.01), (0.03,0.015), (0.04,0.02), and (0.05,0.025).
Thus, we produce 25 datasets as candidates in total.
For playstyle prediction, we also produce the corresponding 25 datasets $T_{s,n}$ as targets. 

In this experiment, we use three different discrete representations from the training of HSD as described in Section \ref{Hierarchical State Discretization}, where the HSD model includes one and only one hierarchy above the base hierarchy, and the discrete state space is set to $2^{20}$.
Table~\ref{Table:TORCS-Style-Accuracy-Baseline} shows a playsytle accuracy, 69.73\%, averaged from the three HSD representations. 
This table also shows the playstyle accuracy for less datasets in the following two experiments. 
First, given a noise level $n$ at random, consider the five speed datasets, $C(s,n)$, where $s$ are the above five speeds. 
Second, similar to the first, given a speed $s$ at random, consider the five noise datasets, $C(s,n)$, where $n$ are the above five noise levels. 
The results shows higher playstyle accuracy, 91.20\% for five speeds, and 81.33\% for five noise levels. 

For comparison, we present the following three different metrics. 
First, if we still use discrete representation, we consider two metrics with different $\phi$. 
The first, called {\em Pixel}, straightforwardly uses the raw pixel values of 4 consecutive screen observations as the states, namely $\overline{s}=\phi(o)=o$.
The second, called {\em LRD}, uses low-resolution downsampling that resizes the raw screens (of observations) from (64,64) to (8,8) with bilinear downsampling and also shrinks the intensity value range from 0-255 to 0-15 by dividing the intensity value by 16 and discarding the remainder.
The last metric, called \textit{Continuous}, does not use discrete representation. 
It simply follows a common image distance metric used in generative models, namely, FID~\citep{FID}, which simply compares the distance of two image datasets in terms of the classification latent distribution, trained from ImageNet~\citep{ImageNet} for identifying the similarity between real images and generative images. 
However, since the model used for FID is for ImageNet and cannot directly be used in our game environment, we use
the latent distribution of our HSD model, as those of $z_{e}^{1}$ in Figure \ref{Figure:HSD}.

For the three metrics, we experiment in the same way as that for HSD. 
Table~\ref{Table:TORCS-Style-Accuracy-Baseline} show that playstyle accuracies for these methods are obviously much lower than that for HSD. 
Note that the performances of Pixel and LRD are low since there are no intersection states in the experiment.
For Continuous, it runs out of time for the one with 25 styles, we can observe that HSD clearly outperforms from five speed styles and noise styles. 
The metric Continuous does not have high accuracy since the observation distribution does not reflect actions which are important in terms of playstyle. 


\begin{table}
    \centering
    \begin{tabular}{llll}
    \toprule
    5 Speed Styles & $2^{16}$ & $2^{20}$ & $2^{24}$ \\
    \midrule
    $t = 1$ & 89.33 & 89.60 & \textbf{91.47}\\
    \midrule
    $t = 2$ & 89.73 & \textbf{91.20} & 89.60 \\
    \midrule
    $t = 4$ & 90.13 & \textbf{90.27} & 86.67 \\
    \bottomrule
    \end{tabular}
    \centering
    \begin{tabular}{llll}
    \toprule
    5 Noise Styles & $2^{16}$ & $2^{20}$ & $2^{24}$ \\
    \midrule
    $t = 1$ & \textbf{74.13} & 72.67 & 73.60 \\
    \midrule
    $t = 2$ & 78.27 & \textbf{81.33} & 78.67 \\
    \midrule
    $t = 4$ & 84.53 & \textbf{85.07} & 82.40 \\
    \bottomrule
    \end{tabular}
    \caption{The comparison of using different size of the HSD state space and threshold in evaluating the playstyle accuracy (\%) in TORCS. Please refer to Section~\ref{Style On Rule-based AI Bot: TORCS} for detailed description.}
    \label{Table:TORCS-Style-Accuracy-Search}
\end{table}

\begin{table}
    \centering
    \begin{tabular}{lllll}
    \toprule
    \quad & Nitro & Surface & Position & Corner \\
    \midrule
    Pixel & 16.67 & 16.67 & 16.67 & 16.67 \\
    \midrule
    LRD & 33.33 & 16.67 & 33.17 & 42.50 \\
    \midrule
    HSD & \textbf{93.11} & \textbf{99.83} & \textbf{99.67} & \textbf{99.56} \\
    \bottomrule
    \end{tabular}
    \caption{RGSK Playstyle Accuracy (\%). Please refer to Section~\ref{Style On Human Player: RGSK} for detailed description.}
    \label{Table:RGSK-Trajectory-Style-Accuracy}
\end{table}

\begin{table*}
    \centering
    \begin{tabular}{lllllllll}
    \toprule
    \quad & Asterix & Breakout & MsPacman & Pong & Qbert & Seaquest & SpaceInvaders \\
    \midrule
    Pixel & 67.10 & 59.05 & 70.95 & 34.20 & 30.00 & 14.95 & 30.35 \\
    \midrule
    LRD & 92.20 & 59.85 & \textbf{96.75} & \textbf{94.15} & 83.70 & 42.70 & 56.10 \\
    \midrule
    HSD & \textbf{98.65} & \textbf{93.83} & 93.85 & 91.55 & \textbf{94.58} & \textbf{93.87} & \textbf{79.72} \\
    \bottomrule
    \end{tabular}
    \caption{Atari Playstyle Accuracy (\%). Please refer to Section~\ref{Style On Learning-based AI Bot: Atari2600} for detailed description.}
    \label{Table:Atari-Model-Accuracy}
\end{table*}

We further study the proper size of state space and the threshold $t$ of our metric in Table~\ref{Table:TORCS-Style-Accuracy-Search}.
It is clear that using a large threshold tends to lead to high accuracy in noise styles, which inherently include big randomness. 
However, the different thresholds do not make large difference in speed styles.
From the table, we observe that $2^{20}$ is an appropriate size of state space since it performs well in most cases.
Thus, in the rest of our experiments we use $t = 2$ and $2^{20}$ state space as a default setting.
More comparisons and experiment details can be found in our appendix.

\subsection{RGSK}
\label{Style On Human Player: RGSK}
In this experiment, we further investigate our playstyle metric with human players, based on another racing game, RGSK, in which its development pack is available on the Unity Asset Store.
For RGSK, we use Unity ML-Agents Toolkit~\citep{Unity-ML-Agents} to sample data, pairs of observations and actions, from human players.
Each of human players is requested to maintain a consistent playstyle during game playing, which is sampled as a dataset representing the playstyle. 
For example, a player are requested to use the $N_{2}O$ boosting system, called Nitro in this paper, to accelerate the car in his/her own style. 
In this experiment, six players, denoted by $p_1$ to $p_6$, are requested to play in their own playstyles of using Nitro, where six datasets $C_1$ to $C_6$ from the six players are collected respectively as candidates. 

Table \ref{Table:RGSK-Trajectory-Style-Accuracy} shows that the playstyle accuracy for HSD reaches 93.11\%, which is much better than those for Pixel and LRD. 
The experiment settings are nearly the same as TORCS, except for the following differences: 
a screen is 72$\times$128, instead of 64$\times$64; the action space is discrete with 27 actions, a combination of three steering directions, three acceleration levels and three driving direction controls (i.e. driving forward, backward, and neutrally). 

In addition to Nitro, we do experiments for other playstyles, where the players are requested to play consistently in their own styles in the following three style dimensions, called Surface, Position and Corner.
For Surface, players are requested to maintain their own playstyles consistently on driving on either the road surface or the grass surface. 
For Position, players are requested to maintain their own playstyles consistently on driving the car in the inner or outer of the track.
For Corner, players are requested to maintain their own playstyles consistently on passing a corner via drifting or slowing down with a break. 
Table \ref{Table:RGSK-Trajectory-Style-Accuracy} also includes the performances of these playstyle dimensions. 
From the table, the playstyle accuracy for HSD reaches above 99\% for these playstyle dimensions, which again shows consistent outperformance in comparison to those for Pixel and LRD. 
The aforementioned experiments also demonstrate the generality of our metric on different playstyle dimensions, as our metric does not need any prior knowledge about which kinds of playstyles to measure.

\subsection{Atari 2600}
\label{Style On Learning-based AI Bot: Atari2600}
Finally, we measure the playstyles of learning-based AI agents for seven Atari games~\citep{atari}. 
These agents are trained by four available RL algorithms, i.e.  DQN~\citep{dqn}, C51~\citep{C51}, Rainbow~\citep{Rainbow}, and IQN~\citep{IQN}), provided by a deep reinforcement learning framework Dopamine~\citep{dopamine,Atari-Zoo}. 
For each of these algorithms, five models are trained from initially different random seeds. 
Thus, there are 20 models in total serving as 20 agents, each of which is presumed to play in a unique playstyle. 

The experiment measures the playstyle accuracy in a similar way to those in the previous sections. 
Namely, 20 agents are used to generate respectively 20 datasets as candidates. 
The experimental settings are similar to those in the previous subsections, except for the following modifications. 
Each screen for observation is $84 \times 84$ grayscale images, and the action space is discrete. 


Table~\ref{Table:Atari-Model-Accuracy} shows the performances of the three playstyle metrics, Pixel, LRD and HSD.
We observe a phenomenon as follows. 
Since observations have low randomness in Atari games, the intersection of states of Atari games for both Pixel and LRD are larger than those in the previous subsections, thus leading to the case that these two metrics have much better performances than those in the previous subsections (i.e. TORCS and RGSK). 
Nevertheless, even under such phenomenon, our playstyle metric HSD remains to outperform the other two for five games, while performing slightly lower than LRD but still quite competitive for the other two games (i.e. MsPacman and Pong). 

\section{Conclusion}
\label{Conclusion}
We propose a novel playstyle metric for video games to evaluate playstyle distances and predict playstyles, based on a novel scheme of learning hierarchical discrete representations. 
To our knowledge, such a metric is the first of its kind without prior knowledge of the games. 
Our experiment results shown in Section \ref{Experiments} also demonstrate high playstyle prediction accuracy of this metric on video game platforms, including TORCS, RGSK,and seven Atari games.  
In fact, our work also leaves some open problems for the future work.
For example, measure the playsytles based on segments of trajectories, not just pairs of observations and actions, investigate the transitivity of playstyle distances, and judge whether an AI acts like a human.
Besides, we make an assumption that each player has a single playstyle for the simplicity of analysis. It is possible to extend our metric to a more general case, which contains more than one playstyle per person/dataset.

\section*{Acknowledgments}
\label{Acknowledgments}
This research is partially supported by the Ministry of Science and Technology (MOST) of Taiwan under Grant Number 109-2634-F-009-019 and 110-2634-F-009-022 through Pervasive Artiﬁcial Intelligence Research (PAIR) Labs, and the computing resource partially supported by National Center for High-performance Computing(NCHC) of Taiwan. The authors also thank anonymous reviewers for their valuable comments.

\begin{contributions} 
    Briefly list author contributions.
    This is a nice way of making clear who did what and to give proper credit.

    H.~Q.~Bovik conceived the idea and wrote the paper.
    Coauthor One created the code.
    Coauthor Two created the figures.
\end{contributions}

\begin{acknowledgements} 
    Briefly acknowledge people and organizations here.

    \emph{All} acknowledgements go in this section.
\end{acknowledgements}

\bibliography{lin_103-supp}

\clearpage
\appendix
\section*{Appendix}
\label{Appendix}

\section{A Study of Some Variants of Playstyle Distance}
In our proposed playstyle metric, we suggest using 2-Wasserstein distance metric in computing the distance of action distributions and using the expected version in gathering the distances over the overlapping states.
Here, we investigate some variants of calculating the playstyle distance on the racing games.
We first give in Subsection~\ref{Unifrom distance and expected distance} the comparison of using a uniform version and expected version as our playstyle distance (cf. Equation (4) and (5) in the main paper); then in Subsection~\ref{Different probability distance metrics}, we conduct the study of using different distances between probability distributions to compute the policy distance.

\subsection{Uniform Distance or Expected Distance}
\label{Unifrom distance and expected distance}
As described in Section \ref{General Playstyle Metric}, we propose to use the uniform version (cf. Equation (4)) or the expected version (cf. Equation (5)) to compute playstyle distances, which are modelled from slightly different aspects. 
In Table~\ref{Table: Uniform vs Expected - TORCS}, we provide the experimental results for the two versions under the setting of using $2^{20}$ state space with different threshold $t$ in $\overline{S_{\phi}'}$, for making comparison on the playstyle accuracy in TORCS, while in  Table~\ref{Table: Uniform vs Expected - RGSK} the results in the RGSK racing game are presented. We can observe that, the expected version is apparently better in TORCS, and the difference between these two versions of playstyle distance is marginal in RGSK.

\subsection{Different Distribution Distance Metrics}
\label{Different probability distance metrics}
In our playstyle metric, we use 2-Wasserstein distance ($W_2$) as the distance metric for action distribution.
For simplicity of discussion, we use $\pi_{A}(\overline{s})$ to denote the action distribution of a dataset $A$ under a specific state $\overline{s}$.
For the policy in the discrete action space, we treat the categorical probability distribution as a simple vector according to the action probability and directly compute the $L_2$ distance between two action vectors as follows.
\begin{equation}
\label{equation:discrete_w2}
W_{2}(\pi_{A},\pi_{B}|\overline{s}) = \|\pi_{A}(\overline{s})-\pi_{B}(\overline{s})\|_{2} \text{,}
\end{equation}
In video game AI, an action-value is usually presented as an action index, and there is no specific relation among these indices in terms of their numerical difference. Hence, the way we present the action distribution still meets the definition in the Wasserstein metric, which indicates that the distances based on each of the action values are of the same importance.
For the policy in the continuous action space, we use the multivariate normal distribution to formulate the action distributions, and the $W_2$ distance is defined by the formula~\citep{W2-Metric} as follows.
\begin{equation}
\label{equation:continuous_w2}
\begin{aligned}
& W_{2}(\pi_{X},\pi_{Y}|\overline{s}) = \|m_{X}(\overline{s})-m_{Y}(\overline{s})\|_{2} + \\ & trace(C_X(\overline{s}) + C_Y(\overline{s}) - 2 (C_Y(\overline{s})^{1/2}C_{X}(\overline{s})C_Y(\overline{s})^{1/2})^{1/2}) \text{,} \\
& \text{where } \pi_{X}(\overline{s})=\mathcalold{N}(m_{X}(\overline{s}),C_{X}(\overline{s})) \text{,} \\
& \text{and } \pi_{Y}(\overline{s})=\mathcalold{N}(m_{Y}(\overline{s}),C_{Y}(\overline{s})) \text{.} \\
\end{aligned}
\end{equation}
$m(\overline{s})$ and $C(\overline{s})$ are the mean vector and the covariance matrix respectively of the multivariate normal distribution under a specific state $\overline{s}$.

In machine learning, Kullback–Leibler divergence (mostly being abbreviated as KL divergence) is a common metric (sometimes, we call it "distance") to measure the difference between two probability distributions.
Its computation upon the discrete probability distributions is as follows:
\begin{equation}
    D_{KL}(P\|Q) = \underset{x \in \mathcalold{X}}{\sum}P(x) \log(\frac{P(x)}{Q(x)})
\end{equation} 
We therefore also experiment to use KL divergence instead of $W_2$ in the RGSK playstyle comparison, which has a discrete action space.
Since KL divergence is an asymmetric metric, we also try its average variant $M_{KL}(P,Q)$ as follows.
\begin{equation}
    M_{KL}(P,Q) = \frac{D_{KL}(P\|Q) + D_{KL}(Q\|P)}{2}
\end{equation} 
In addition to $W_2$ and KL divergence, we also include $W_1$ metric, which uses $L_1$ norm in Wasserstein metric instead of $L_2$ norm, in our comparison.
As the results based on all the aforementioned metrics (i.e. $W_2$, KL divergence, and $W_1$) being shown In Table~\ref{Table: W1 vs W2 vs LK vs AvgKL - RGSK}, where the state space is $2^{20}$ with the threshold $t = 2$, we observe that using Wasserstein metric has better performance than KL divergence, and $W_1$ is sightly better than $W_2$ in this case. Please note that, even though the better performance of $W_1$, we adopt the $W_2$ as the distribution distance used for our playstyle metric computation due to its common application as well as the capability of reflecting the covariance between different action dimensions.

\begin{table}
    \centering
    \begin{tabular}{lll}
    \toprule
    \quad  & 5 Speed Styles & 5 Noise Styles \\
    \midrule
    Uniform ($t = 2$) & 76.00 & 65.20\\
    \midrule
    Uniform ($t = 4$) & 83.47 & 74.40\\
    \midrule
    Expected ($t = 2$) & 91.20 & 81.33\\
    \midrule
    Expected ($t = 4$) & 90.27 & 85.07\\
    \bottomrule
    \end{tabular}
    \caption{The comparison of TORCS playstyle accuracy (\%) between using the uniform version (cf. Equation (4)) or the expected version (cf. Equation (5)) as our playstyle distance.}
    \label{Table: Uniform vs Expected - TORCS}
\end{table}

\begin{table}
    \centering
    \begin{tabular}{lllll}
    \toprule
    \quad & Nitro & Surface & Position & Corner \\
    \midrule
    Uniform ($t =2$) & 92.33 & 99.28 & 99.33 & 99.50\\
    \midrule
    Uniform ($t = 4$) & 86.94 & 96.44 & 95.56 & 95.61\\
    \midrule
    Expected ($t = 2$) & 93.11 & 99.83 & 99.67 & 99.56\\
    \midrule
    Expected ($t = 4$) & 88.78 & 98.39 & 98.33 & 98.44\\
    \bottomrule
    \end{tabular}
    \caption{The comparison of RGSK playstyle accuracy (\%) between using the uniform version (cf. Equation (4)) or the expected version (cf. Equation (5)) as our playstyle distance.}
\label{Table: Uniform vs Expected - RGSK}
\end{table}

\begin{table}
    \centering
    \begin{tabular}{lllll}
    \toprule
    \quad & $W_1$ & $W_2$ & $D_{KL}$ & $M_{KL}$ \\
    \midrule
    Nitro & \textbf{95.28} & 93.11 & 37.44 & 76.78\\
    \midrule
    Surface & \textbf{100.00} & 99.83 & 74.28 & 96.06\\
    \midrule
    Position & \textbf{99.83} & 99.67 & 58.06 & 91.94\\
    \midrule
    Corner & \textbf{99.83} & 99.56 & 56.33 & 87.33\\
    \bottomrule
    \end{tabular}
    \caption{The comparison in terms of playstyle accuracy (\%) among adopting different metrics for the action distributions. The video game environment used in this experiment is based on RGSK.}
    \label{Table: W1 vs W2 vs LK vs AvgKL - RGSK}
\end{table}

\begin{table*}
\begin{subtable}{1\textwidth}
    \centering
    \begin{tabular}{lllllllll}
    \toprule
    \quad & $2^{0}$ & $2^{4}$ & $2^{8}$ & $2^{16}$ & $2^{32}$ & $2^{64}$ & Pixel & VQ-VAE\\
    \midrule
    $t = 2$ & 79.40 & 70.93 & 68.93 & \textbf{89.73} & 84.53 & 66.40 & 20.00 & 20.00 \\
    \midrule
    $t = 4$ & 79.40 & 68.53 & 72.80 & \textbf{90.13} & 88.80 & 43.87 & 20.00 & 20.00 \\
    \bottomrule
    \end{tabular}
    \caption{The playstyle accuracy(\%) over 5 different target speeds.}
\end{subtable}

\begin{subtable}{1\textwidth}
    \centering
    \begin{tabular}{lllllllll}
    \toprule
    \quad & $2^{0}$ & $2^{4}$ & $2^{8}$ & $2^{16}$ & $2^{32}$ & $2^{64}$ & Pixel & VQ-VAE\\
    \midrule
    $t = 2$ & 38.28 & 52.80 & 63.87 & \textbf{78.27} & 74.53 & 55.73 & 20.00 & 20.00 \\
    \midrule
    $t = 4$ & 38.28 & 53.07 & 61.73 & \textbf{84.53} & 81.33 & 41.33 & 20.00 & 20.00 \\
    \bottomrule
    \end{tabular}
    \caption{The playstyle accuracy(\%) over 5 different noise levels.}
\end{subtable}
\caption{The experiments of using different sizes of the state space in TORCS.}
\label{Table:TORCS-Style-Accuracy-Different-Threshold}
\end{table*}

\section{An Example of Calculating the Playstyle Distance}
\begin{figure*}
    \begin{center}
        \includegraphics[width=1\textwidth]{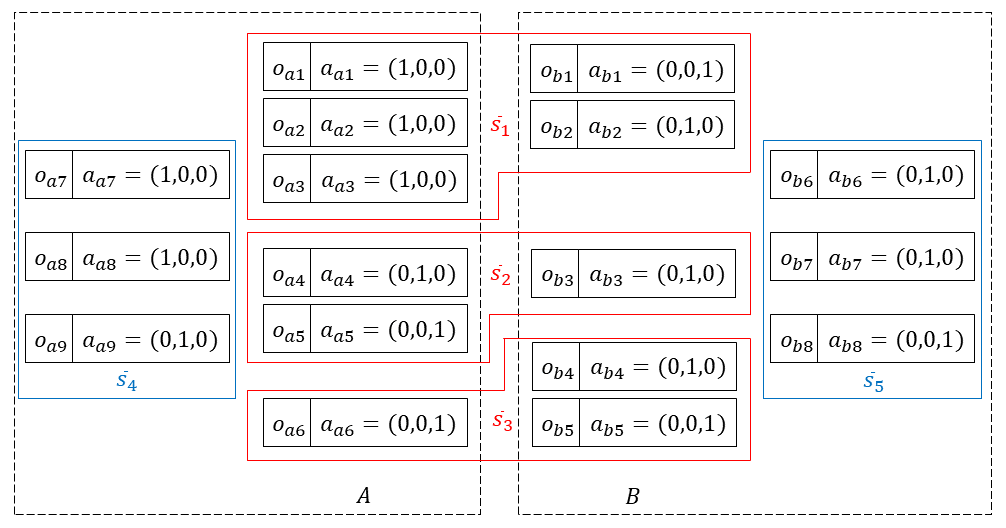}
    \end{center}
    \caption{An example scenario for showcasing the computation of playstyle distance.}
    \label{Figure:Policy Distance Example}
\end{figure*}
We give an example for calculating the playstyle distance, with the scenario as shown in Figure~\ref{Figure:Policy Distance Example}, where the action space is discrete and we adopt the $W_2$ distance as the distribution metric.

First, we have to compute the policy distances over intersection states (in the red border regions) as follows.
\begin{equation}
\begin{aligned}
&\overline{d}_{\phi}(A,B|\overline{s_1}) = \|(1,0,0) - (0,0.5,0.5)\|_2 \approx 1.225 \text{, }  \\
&\overline{d}_{\phi}(A,B|\overline{s_2}) =  \|(0,0.5,0.5) - (0,1,0)\|_2 \approx 0.707 \text{, } \\
&\overline{d}_{\phi}(A,B|\overline{s_3}) =  \|(0,0,1) - (0,0.5,0.5)\|_2 \approx 0.707 \text{.}
\end{aligned}
\end{equation}
Next, we can calculate the playstyle distance as following the Equation (4):
\begin{equation} 
\begin{split}
    d_\phi(A,B) & = \frac{d_{\phi}(A|B)}{2} + \frac{d_{\phi}(B|A)}{2} \\
    & \approx \frac{1.225 \times 0.5 + 0.707 \times 0.333 + 0.707 \times 0.167}{2} \\ & + \frac{1.225 \times 0.4 + 0.707 \times 0.2 + 0.707 \times 0.4}{2} \\
    & \approx \frac{0.966}{2} + \frac{0.914}{2} \\
    & = 0.940
\end{split}
\end{equation}
Finally, when we use a state count threshold $t = 2$ in the policy distance, $\overline{s_2}$ and $\overline{s_3}$ are then not included in the intersection state $\overline{S'_\phi}$, in which it leads to the playstyle distance computed as follows:
\begin{equation} 
\begin{split}
    d_\phi(A,B) & = \frac{d_{\phi}(A|B)}{2} + \frac{d_{\phi}(B|A)}{2} \\
    & \approx \frac{1.225 \times 1.0}{2} + \frac{1.225 \times 1.0}{2} \\
    & = 1.225
\end{split}
\end{equation}

\section{A Study of Selecting the Size of State Space}
How to select a proper state space in our metric is an important topic.
We present a wide search of using different sizes of the state space in TORCS, where the results are shown in 
Table~\ref{Table:TORCS-Style-Accuracy-Different-Threshold} based on the playstyle accuracy. We can observe that:
If the state space is too small, it is hard to obtain good accuracy; However, if we use a large state space like $2^{64}$, the inherent problem of high-dimensional observations would occurs, in which the size of the intersection state set is too small to give a precise playstyle computation. In results, 
$2^{16}$ $\sim$ $2^{32}$ is a good range to explore, where we suggest to use $2^{20}$ as a default space size in our main paper. 
Also, a vanilla VQ-VAE has a too large state space that can not find intersection states.

\section{Experimental Settings}
\begin{figure*}
    \begin{center}
        \includegraphics[width=1\textwidth]{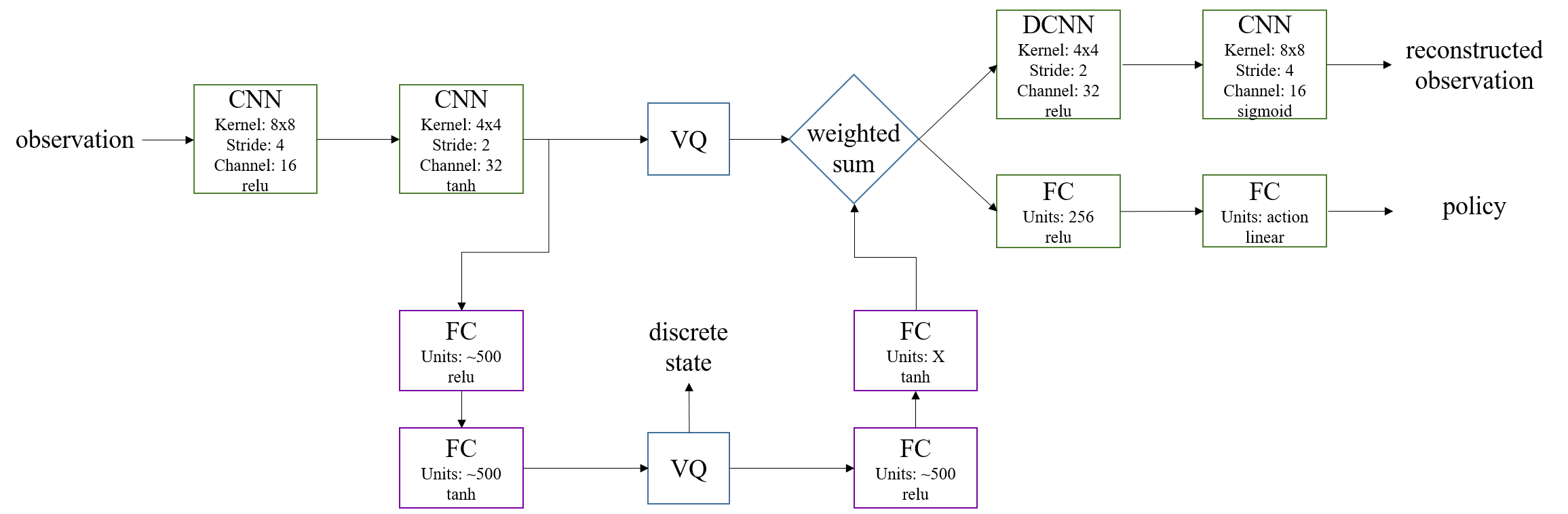}
    \end{center}
    \caption{The architecture of the HSD model used in all video game experiments.}
    \label{Figure:Video Game HSD}
\end{figure*}

In our experiments, we use the same network architecture in all video games, as described in the following. The encoder is based on a simple version of IMPALA network, which is proposed by \citet{IMPALA}.
We show the full architecture in Figure~\ref{Figure:Video Game HSD}, where the unit count $X$ in the hierarchical decoder depends on the dimension of the CNN encoder output. For example, in Atari games, $X=11 \times 11 \times 32$. 
The unit count around 500 means that we use the closest divisible number to the number of cell size $B$. For example, in $2^{16}$ state space, the number is 512, which is divisible to 16, and in $2^{20}$ state space, the number is 500, which is divisible to 20.
A truncated normal initializer initializes the embedding tensors with a standard deviation of 0.02.
When training these HSD models, we add noises sampled from a normal distribution with zero mean and standard deviation 4 to the raw pixel values (0-255); after that, the pixel values are divided by 255 as a simple method for pre-processing the observations.

\subsection{Training HSD models in different experiments}
The training data for the HSD models in TORCS experiments is sampled from human players with a joystick. There are 14502 observation-action pairs in the training data; For training the HSD models in RGSK experiments, we sample some demonstrations from human players. There are 26 episodes in the training data, with each consisting of two complete circuits; For training the HSD models in Atari experiments, we use the first Rainbow~\citep{Rainbow} model in Dopamine~\citep{dopamine} with $\epsilon$-greedy ($\epsilon$ = 0.01) to sample 100 episodes as the training data for HSD models in each game.

\end{document}